\documentclass[11pt]{article}
\pdfoutput=1
\usepackage{amsmath, amssymb, amscd, amsthm, amsfonts}
\usepackage{graphicx}
\usepackage{hyperref}

\usepackage{amssymb,amsmath,amsthm,latexsym,amscd,amsfonts,mathrsfs}
\usepackage[T1]{fontenc}
\usepackage[utf8]{inputenc}
\usepackage{authblk}
\usepackage{graphics}
\usepackage{tikz}
\usetikzlibrary{calc,plotmarks}
\usetikzlibrary{arrows}
\usepackage{cite}
\usepackage{epsfig}
\usepackage{hyperref, comment, cite}
\usepackage{epstopdf}
\usepackage{tikz}
\usepackage{bbm}
\usepackage[justification=centering]{caption}
\usepackage{dsfont}
\usepackage{setspace}
\usepackage{float}
\usepackage{psfrag}
\usepackage{relsize}
\usepackage{etoolbox}
\usepackage{bbm}
\usepackage{hyperref}
\usepackage[ruled,vlined]{algorithm2e}

\newtheorem{mydef}{Definition}
\newtheorem{theorem}{Theorem}

\newtheorem{lemma}{Lemma}

\def\VR{\kern-\arraycolsep\strut\vrule &\kern-\arraycolsep}
\def\vr{\kern-\arraycolsep & \kern-\arraycolsep}

\newcommand{\ceil}[1]{\left\lceil #1 \right\rceil}

\newcommand{\Ac}{\mathcal{A}}

\newcommand{\Dc}{\mathcal{D}}

\newcommand{\Fc}{\mathcal{F}}

\newcommand{\Pc}{\mathcal{P}}

\newcommand{\Ncal}{\mathcal{N}}
\newcommand{\Pcal}{\mathcal{P}}

\newcommand{\Pcb}{\boldsymbol{\mathcal{P}}}

\newcommand{\sumin}{\sum_{i=1}^n}

\newcommand{\onek}{\{1,2,\ldots,k\}}

\newcommand{\Sigmah}{{\hat{\Sigma}}}

\newcommand{\R}{\mathbb{R}}
\newcommand{\E}{\mathbb{E}}

\def\eps{\epsilon}

\DeclareMathOperator\C{C}

\def\textiid{i.i.d.\@\xspace}
\newcommand\iid{\ifmmode\text{ i.i.d. } \else \textiid \fi}

\newcommand{\ind}{\mathbbm{1}}

\newcommand{\beqs}{\begin{equation*}}
\newcommand{\eeqs}{\end{equation*}}
\newcommand{\beq}{\begin{equation}}
\newcommand{\eeq}{\end{equation}}
\DeclareMathOperator*{\argmax}{arg\,max}
\DeclareMathOperator*{\argmin}{arg\,min}

\newcommand{\I}{I}

\newcommand{\Prob}{\mathbb{P}}

\newcommand{\rtht}{R_{\tilde{\theta}}}
\newcommand{\RI}{R_I}
\newcommand{\tht}{\tilde{\theta}}

\newcommand{\ksig}{\Gamma}

\title{Sequential Estimation under Multiple Resources: \\ a Bandit Point of View
}

\author[]{Alireza Masoumian\thanks{alireza.masoumian@ee.sharif.edu}}
\author[]{Shayan Kiyani\thanks{kiyani.shayan@ee.sharif.edu}}
\author[]{Mohammad Hossein Yassaee\thanks{yassaee@sharif.edu}}
\affil[]{Sharif University of Technology}
\date{Sep 2021}

\begin{document}

\maketitle

\begin{abstract}
The problem of Sequential Estimation under Multiple Resources (SEMR) is defined in a federated setting. SEMR could be considered as the intersection of statistical estimation and bandit theory. In this problem, an agent is confronting with k resources to estimate a parameter $\theta$. The agent should continuously learn the quality of the resources by wisely choosing them and at the end, proposes an estimator based on the collected data.
In this paper, we assume that the resources' distributions are Gaussian. The quality of the final estimator is evaluated by its mean squared error. Also, we restrict our class of estimators to unbiased estimators in order to define a meaningful notion of regret. The regret measures the performance of the agent by the variance of the final estimator in comparison to the optimal variance. We propose a lower bound to determine the fundamental limit of the setting even in the case that the distributions are not Gaussian. Also, we offer an order-optimal algorithm to achieve this lower bound.
\end{abstract}

\section{Introduction}\label{section-introduction}
Datasets are constantly growing in size and complexity. This leads to several problems in statistical estimation and inference. The previous well-known algorithms that performed well on small datasets are now facing severe computational challenges
\cite{fan2014challenges}
\cite{franke2016statistical}
\cite{wang2016statistical}. These computational burdens, together with other issues such as privacy\cite{smith2012big}, motivated a long line of research on the distributed estimation in the past years. An illustrating example is massive sensor networks. Assume we are interested in measuring a feature of an environment. To do so, we have access to several sensors spread in that environment. Now the question is how should we cope with this situation. In other words, how should we clarify which sensors are more valuable for our purpose and which ones we should exploit on? These are some frequently appeared questions on a variety of real-world applications\cite{shang2015probabilistic} \cite{jardak2010parallel}.
In this paper, we try to build new insight into the problem of sequential distributed estimation. To be more specific, we attempt to establish a connection between the problem of distributed estimation and bandit theory. From bandit theory's point of view, our work is trying to model how a rational agent should act while she is facing distributed estimation problem over time. Even though our platform is closely related to the conventional settings in distributed estimation and bandit theory, it has some fundamental differences with both of them that arise several challenges.

\subsection{Related Works}
Although we could not find a work precisely aligned with the direction of our research, there are some related works in both areas of bandit theory and distributed estimation.
\subsubsection{Bandit}
In the past years, Bandit theory drew lots of attention from a variety of areas, such as computer science, statistics, operation research, and economics. It has been studied since the pioneering work of Thompson \cite{thompson1933likelihood} with a big bulk of research in the last 20 years \cite{hao2020adaptive},\cite{agrawal2013thompson},\cite{chen2021statistical}. Our work essentially relates to the stochastic multi-armed bandit problem. The closest works to our framework are the ones in which variance of the underlying distributions of arms are considered
\cite{audibert2009exploration}
\cite{borsos2018online}
\cite{audibert2007variance}. However, we work with a novel notion of regret that focuses on the variances of distinct arms directly. To be more specific, the samples that the learner obtains from the arms do not explicitly appear in the notion of regret which is not the case of conventional settings in multi-armed bandit. In the stochastic bandit, the usual notion of regret is based on an aggregation of obtained rewards, the value of driven data, but in our case, the samples contribute to the regret after some processing. Therefore, this problem needs some additional effort to be modeled in the bandit environment.
\subsubsection{Distributed Estimation}
On the other hand, there was a big surge of activity on the problem of distributed estimation in recent years. In this setting, the central fusion interacts with a couple of data collectors under some limitations. The aim is to estimate an underlying parameter by  wisely adopting a policy for transferring  data and using them at the fusion center. A variety of limitations were considered such as, communication constraints in the number of transferred bits between data collectors and the fusion center \cite{barnes2020lower},\cite{han2018geometric},\cite{han2018distributed},\cite{zhang2013information},\cite{braverman2016communication} or communication through a differential private channel \cite{barnes2020fisher}. In some cases with particular distribution families, the effects of these limitations have been clarified, there are some lower bounds for the best possible performance under these constraints
\cite{zhang2013information}
\cite{acharya2021interactive}
\cite{barnes2020lower}
 and their corresponding achieving policies
\cite{garg2014communication}
\cite{han2018geometric}
\cite{han2018distributed}.\\
There are some similarities in the setting of our interest and the distributed estimation, but the differences between the problem on hand and the distributed estimation are conspicuous. First, In our problem, the fusion center picks a data collector in each round intentionally, which is not the case in the distributed estimation. Second, there is no implicit structure in the communication protocols of the distributed estimation.
In more detail, data collectors can send even a single bit over the link about their data or some encoded version of that in the distributed estimation setting. On the other hand, In this setting, they should send a sample of data entirely. As a result,  the known theorems in the distributed estimation literature are not directly applicable to our setting. These differences motivated us to use another approach that comes from bandit theory instead of previously developed arguments in the federated learning line of research.

\subsection{Main Result}
Our main contribution is in modeling the problem of distributed estimation as a multi-armed bandit problem. Through achieving the final objective, estimating the common mean of the arms' distributions, we have to optimize two sub-objectives. First, finding a comprehensive policy for the arm selecting to drive data from the arms with the smallest variances. Second, designing an estimator to estimate the parameter based on obtained samples. We offer a notion of regret which precisely measures the performance of the entire Algorithm in both sub-objectives. In this work, we gained from a UCB-based algorithm for the arm selecting part. Also for the second part, estimation of the parameter, the best estimator in the $l_2$ norm is clearly determined if we restrict the arms' distributions to the Normal distributions. In the end, we put up a lower bound on the performance of any algorithm in this setting without any restriction in the distribution family of the data. This confirms the optimality of our algorithm up to a logarithmic factor.
\subsection{Organization of The Paper}
In section \ref{Section2} we briefly present the background that is required for reading what follows and also the formal definition of the investigated problem.  In section \ref{section3} we present our algorithm. In section \ref{section4} we try to give a better illustration of the notion of regret. In section \ref{section5} we drive a lower bound on the performance of any algorithm. Finally, in section \ref{section6} we offer some of the possible ways for researching in the future. The rest of this article would be Acknowledgment, References, and Appendix.
\section{Backgrounds and Problem Definition} \label{Section2}
\subsection{Bandit}
A bandit problem is trying to mathematically model a series of interactions between a learner and an environment. This problem is studied in two main settings, stochastic and adversarial. Our work in this article has a close connection with the stochastic setting of the bandit theory. In the stochastic bandits, the environment is modeled by a finitely many armed machine and each arm is modeled by a probability distribution. The learner in each round chooses an arm and the environment gives him a sample from the arm's distribution as the reward. The rewards of each arm would be an i.i.d samples from its probability distribution over time. The purpose of the learner is to minimize her regret. The regret definition in the bandit theory is a critical argument and different definitions of regret could lead to a variety of problems and the corresponding algorithms for controlling their regrets. Although there is a common intuition in all forms of regrets which is trying to represent the difference of the learner's performance versus "the best performance". Of course how we define this "the best performance", would lead to different regret definitions.

\subsection{Fisher Information}
Let $\{ P_\theta \}_{\theta \in \Theta}$ be a family of probability distributions. Suppose that each $P_\theta$ has density $f(x;\theta)$. For a random variable $ X \sim f(x;\theta)$, we define $l_\theta(x) = \log f(x;\theta) $ as the log-likelihood function and 
$$
l'_\theta(x) = \frac{\partial}{\partial \theta}\log f(x;\theta) = \frac{f'(x;\theta)}{f(x;\theta)}
$$
as the score function. The Fisher information of $\theta$ contained in the random variable $X$, is defined as follows,
$$
\I_\theta(X) = Var_\theta(l'_\theta(x)) = \E_\theta[l'_\theta(x)^2] =\int l'_\theta(x)^2 \  f(x;\theta) \  dx
= \E_\theta\bigg[\frac{\partial^2}{\partial \theta^2}l_\theta(x)\bigg] 
$$
where it could be easily verified that these different forms of Fisher information are equivalent.

Fisher information has a meaningful role in a well-known inequality called Crámer-Rao which states that each unbiased estimators of $\theta$ from samples of $X$, has an inevitable variance. In more details,
$$
Var(\hat\theta) \geq \frac{1}{\I_\theta(X)} \quad \text { for all unbiased estimators } \hat\theta.
$$
and in a multi-dimension version of Crámer-Rao we have,
$$
\Sigma _{\hat\theta} = \E_\theta[(\hat\theta - \E[\hat\theta]) . (\hat\theta - \E[\hat\theta])^T] \succcurlyeq I_\theta(X)^{-1}  \quad \text { for all unbiased estimators } \hat\theta.
$$
where $\Sigma _{\hat\theta}$ is the covariance matrix of $\hat\theta$. Note that the both matrices $\Sigma _{\hat\theta}$ and $\I_\theta(X)^{-1}$ are positive semi definite, hence the comparison of them is meaningful.

\subsection{Problem Definition}
\begin{figure}[h] 
    \centering
    \includegraphics[width=.65\textwidth]{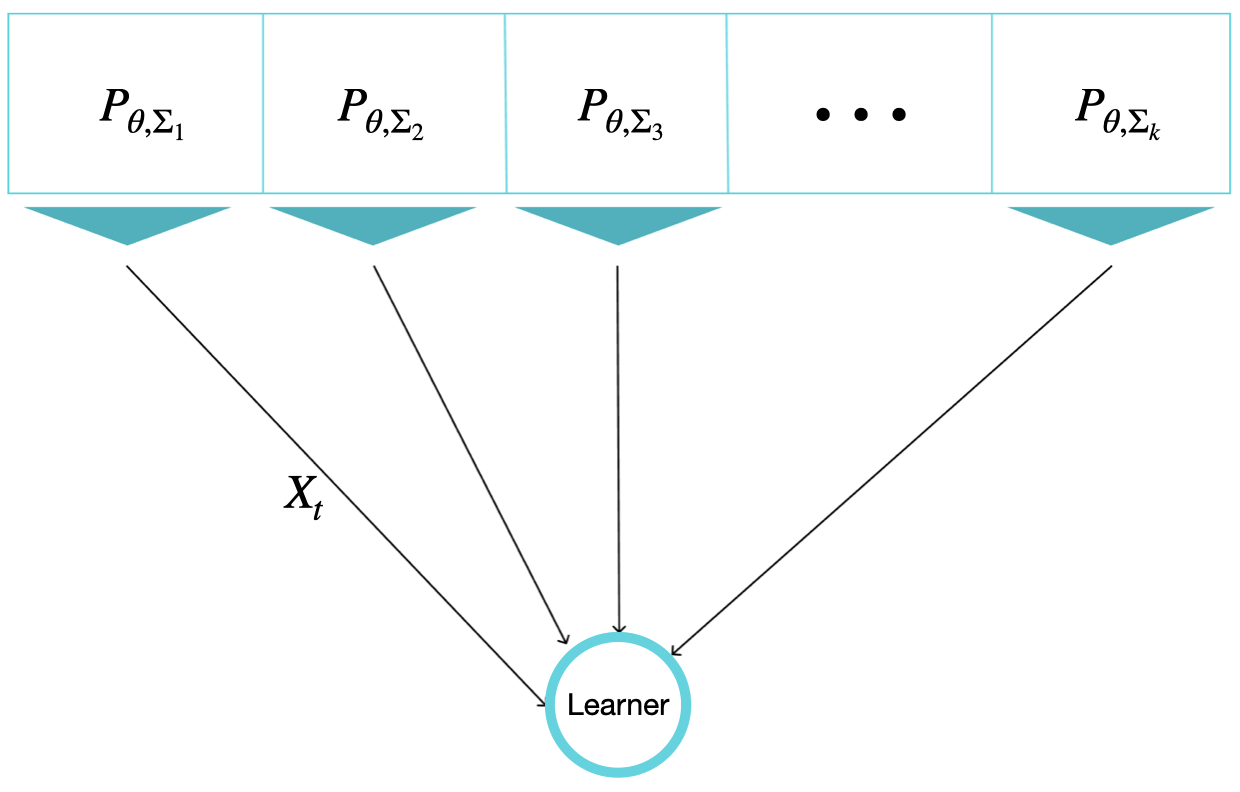}
    \caption{Setting of the problem} \label{Prfigure}
\end{figure}

A SEMR problem is a sequential game between a learner and an environment. The game is played over $n$ rounds, where $n$ is a positive integer number, called the horizon. The environment contains $k$ stochastic sources of data. In each round $t \in [n] $, the learner first chooses a source $A_t$ and then the environment reveals a sampled data $X_t \in \R^d$ from that source. See figure (\ref{Prfigure}).
The main goal of our problem is to estimate an unknown parameter $\theta$ by proper chooses of the sources over time. For the  $i$'th source, we assume a probability distribution $P_{\theta , \Sigma_i}$ belongs to a class of distributions $\mathcal{P}_{\boldsymbol\Theta , \boldsymbol\Sigma}$, in which each distribution could be parametrized with two parameters $\theta \in \boldsymbol\Theta$ and $\Sigma \in \boldsymbol\Sigma $. Note that all sources have the same $\theta$, but, in contrast, the $\Sigma_i$ is different for each source. The learner has no knowledge about both $\theta$ and $\Sigma_i$s. The difference of the $\Sigma_i$s causes the importance of finding the best source to collect our data from it.

Obviously the learner is unaware about the future when choosing her actions, so the selected source $A_t$ should only depend on the history $H_{t-1}=(A_1,X_1,\dots,A_{t-1},X_{t-1})$.
She picks a source $A_t$ with respect to a policy $\pi_t$ which is a mapping from history to the sources. Also, $\hat\theta$ is the final unbiased estimator of the parameter $\theta$ after $n$ steps.

In order to measure the performance of the policy, we define regret as follows,
\begin{align} \label{regret}
Reg(n) = \E \Vert \hat\theta - \theta \Vert^2_2 - Tr((n I_{a^*}^\theta)^{-1}) =  Tr(\E[\tht \tht^T]) - Tr((n I_{a^*}^\theta)^{-1})
\end{align}
where $\tilde\theta = \hat\theta - \theta$ is the normalized estimator of the parameter $\theta$ at the end, and $I_*^\theta$ is the Fisher information matrix with the greatest trace, over all arms, i.e.,
\begin{align}
  a^* = \argmax_{j \in [k]} Tr(I_j^\theta)
\end{align}
When we restrict the class of estimators to unbiased estimators, this regret is a meaningful statement. In the scalar case, It is the difference between the variance of the final estimator and the asymptotically optimal variance. And in the multi-dimensional case it has a same meaning by considering a quantity of covariance matrix. Here we contemplate the trace of matrices, which has a meaningful interpretation related to Mean Squared Error (MSE).\\
The core question in this work is to understand the growth rate of the regret as $n$ grows. In this direction, we presented a novel algorithm in data collecting and a converse bound on regret.
\section{Algorithm} \label{section3}
In section 1, we proposed the problem in general form. In this section, We restrict our distributions such that $\Pcal_{ \boldsymbol\Theta,  \boldsymbol\Sigma} = \{ \Ncal (\theta , \Sigma) | \theta \in \R^d , \Sigma \in \R^{d\times d} \}$. Furthermore,
assume an upper-bound over the Frobinous norm of covariance matrices.
\begin{align} \label{GammaAssumption}
    \forall i \in \onek : \quad \Vert\Sigma_i\Vert_F \leq \ksig
\end{align}
This is an intuitive assumption because we know that the Fisher information matrix of the Gaussian distribution is $\Sigma^{-1}$ and this upper bound in a norm of $\Sigma$ guarantees that all of the arms have enough information about $\theta$.
\subsection{Lower Confidence Bound (LCB)}
Now we propose a UCB-based algorithm. In each round, we compute an index for each arm and then choose the arm with the smallest index.\\
Our index function is as follows,
\begin{align}
\operatorname{LCB}_{i}(t-1, \delta)=\left\{\begin{array}{ll}-\infty & \text { if } T_{i}(t-1)=0 \\ \hat{tr}_{i}(T_i(
t-1))-\sqrt{\frac{8 \log (2 / \delta)}{T_{i}(t-1)}}\Gamma & \text { otherwise }\end{array}\right.    
\end{align}\\
and we have,
\begin{align}
    \bar{X}_i(t-1) = \frac{\sum_{j = 1}^{T_i(t-1)} X_j^i}{T_i(t-1)}
\end{align}
\begin{align}
    \hat{tr}_{i}(T_i(t-1)) = Tr(\Sigmah_i) = Tr(\sum_{j = 1}^{T_i(t-1)} \frac{1}{T_i(t-1)}(X_j^i-\bar{X}_i(t-1))(X_j^i-\bar{X}_i(t-1))^T) 
\end{align}
is our estimation of the covariance matrix of the $i$'th arm and $X_j^i$ is the $j$'th sample of the $i$'th arm.
\begin{algorithm}
\SetAlgoLined
 \KwIn{n, $\Gamma$}
 $t=0$, $\delta=\frac{2}{ n^{\max\{(d-1)^2,2\}}}$, for every $i$ $\in \onek$ : $T_i(0)=0$,\\
 \vspace{3mm}
 $\operatorname{LCB}_{i}(t-1, \delta)=\left\{\begin{array}{ll}-\infty & \text { if } T_{i}(t-1)=0 \\ \hat{tr}_{i}(T_i(t-1))-\sqrt{\frac{8 \log (2 / \delta)}{T_{i}(t-1)}}\Gamma & \text { otherwise }\end{array}\right.$\\
 \While{$t\ \leq n$}{
  $t=t+1$ \\
  $A_t=\arg\min_{i\in \onek}\big(LCB_i(T_i(t),\delta)\big)$\\
  \text{Sample $X_t\ $ from source $A_t$.}\\
  $T_{A_t}(t)=T_{A_t}(t-1)+1$\\
  \text{Calculate} $LCB_{A_t}(T_{A_t}(t),\delta)$.\\
  \text{Calculate} $\hat{tr}_{A_t}(t)$.
 }
\KwOut{$\hat\theta = \frac{1}{n} \sumin X_i$}
\caption{\text{LCB}}
\end{algorithm}
\subsection{Regret Analysis}
Under Gaussian assumption the regret (\ref{regret}) reduces to,
\begin{align}
  Reg(n) = Tr(\E[\tht \tht^T]) - Tr((n I_{a^*}^\theta)^{-1}) = Tr(\E[\tht \tht^T]) - \frac{1}{n} Tr(\Sigma^*) 
\end{align}
Also, by doing some math we have $Tr(\E[\tht \tht^T]) = \frac{1}{n^2} \sum_{t=1}^n Tr(\E\Sigma_{a_t})$ and it implies that, 
\begin{align}
Reg(n) = \frac{1}{n^2} \sum_{t=1}^n Tr(\E\Sigma_{a_t}) - \frac{1}{n} Tr(\Sigma^*) = \frac{1}{n^2} \big [ \sum_{t=1}^n (Tr(\E\Sigma_{a_t}) - Tr(\Sigma^*) ) \big]
\end{align}
Now we define the optimality gap for each arm as $\Delta_i = Tr(\Sigma_{i}) - Tr(\Sigma^*)$ and the counter function of arms as $T(i) = \sum_{t=1}^n \ind\{a_t = i\}$. We assume that the best arm is the first one for convenient, i.e. $\Sigma^* = \Sigma_1$.\\
Now we can rewrite the regret as the final notion,
\begin{align}\label{additive}
    Reg(n) = \frac{1}{n^2}\sum_{i = 1}^k \Delta_i \E[T_i(n)]
\end{align}\\
in this notion our aim reduces to controlling $T(i) \quad \forall i \neq 1$. 
\begin{theorem}\label{order}
The LCB algorithm guarantees the following regret bound,
\begin{align} \label{regretbound}
 Reg(n) \in O(n^{-\frac{3}{2}}\sqrt{\log(n)})
\end{align}
\end{theorem}

\proof
consider equation (\ref{additive}) for the regret. We state that $    \E [T_i(n)] \leq  \frac{ C_d \log(n)}{\Delta_i^2} + 5$ in lemma (\ref{Goodevent}). We separate the range of $\Delta_i$ with a threshold $\Delta_*$, Therefore we have,
\begin{align}
       n^2 Reg(n) = \sum_{i = 1}^k \Delta_i \E[T_i(n)] & =  \sum_{\Delta_i < \Delta_*} \Delta_i \E[T_i(n)] + \sum_{\Delta_i \geq \Delta_*} \Delta_i \E[T_i(n)]  \\ &\leq
        n \Delta_* 
        + k \frac{C_d \log(n)}{\Delta_*} + \sum_{\Delta_i \geq \Delta_*} 5 \Delta_i 
\end{align}
Note that the inequalities are from $\sum_{\Delta_i \geq \Delta_*} \E[T_i(n)] \leq n $ and $\vert\{i : \Delta_i \geq \Delta_*\}\vert < k$.
Finally by choosing the threshold $\Delta_* = \sqrt{\frac{ k \log(n) \C_d}{n}}$ we have,
\begin{align} \label{ineq13}
    Reg(n) \leq \frac{5\sum_{i=1}^{k} \Delta_i }{n^2} +  4 \frac{\ksig \Big(1+\sqrt{\max\{2,(d-1)^2\}}\Big) \sqrt{ 2 k \log(n)} }{n\sqrt{n}} 
\end{align}
The result (\ref{ineq13}) shows the dependency of the algorithm to dimension of distributions. In other words, the upper-bound grows linearly with respect to $d$.  

\section{Regret Decomposition} \label{section4}
From a high-level perspective, our problem contains two components. The first one is an approach to get familiar with the environment. In other words, how we interact with the sources and collecting more informative data is the principal concern of this phase. In the second phase, all that matters is estimating the parameter of interest based on the data collected in the previous phase.
Now in this section,  We offer two different decompositions of the notion of regret. These decompositions illustrate that how each of these phases participates in our final regret separately. Besides, in both of them, we gained from the additivity property of Fisher information. This is because the samples are drawn independently from the distributions.
The first one in section 4.1 originates from the intuition behind the additivity of regret in bandit theory. The second one comes from the foreknowledge behind the statistical properties of Fisher information. In a nutshell, both of these decompositions evaluate the quality of the estimator independently from the quality of the data collection. 

\subsection{First Decomposition}
 As the first decomposition, We can rewrite regret (\ref{regret}) in this way,
\begin{align}
Reg(n) = Tr\big(\E(\tht \tht^T)\big) - Tr\big((n I_{a^*})^{-1}\big) &= Tr\big(\E(\tht \tht^T)\big) - \frac{1}{n^2}Tr\big(\big[\E[\sum_{t=1}^n I_{a_t}^{-1}]\big]\big) \\
 &+ \frac{1}{n^2}Tr\big(\big[\E[\sum_{t=1}^n I_{a_t}^{-1}]\big]\big) - Tr\big((n I_{a^*})^{-1}\big)
\end{align}
The key point in this decomposition is in its second part. This part precisely could be considered as a standard additive regret. In other words, the reward is the value of the Fisher information. Note that, as we mentioned before, this regret is not sensible.
Besides, the first part evaluates the performance of the estimator. For example, in the Algorithm section, we saw that this part is zero in the case of Gaussian distributions. Therefore regret is reduced to the second part which is analyzed earlier.
\subsection{Second Decomposition} \label{secdec} \label{section4.2}
As the second decomposition, We have,
\begin{align} \label{dec2}
Reg(n)= Tr\big(\E(\tht \tht^T)\big) - Tr\big((n I_{a^*})^{-1}\big) = \rtht(n) + \RI(n) 
\end{align}
and we have,
\begin{align}
    \rtht(n) = Tr\big(\E(\tht \tht^T)\big) -  Tr\big(\big[\E[\sum_{t=1}^n I_{a_t}]\big]^{-1}\big) \\
    \RI(n) = Tr\big(\big[\E[\sum_{t=1}^n I_{a_t}]\big]^{-1}\big) -Tr\big((n I_{a^*})^{-1}\big)
\end{align}
This decomposition values the asymptotic properties of Fisher information. The first term represents the gap between the variance of the estimator and the least possible variance among unbiased estimators. This comes from the well-known Cramer-Rao inequality. The equality (\ref{dec2}) will be used in the next section to come up with a lower bound on the performance of any algorithm that benefits from an unbiased estimator.
\section{Lower Bound} \label{section5}
In section (\ref{secdec}), we decomposed our notion of regret to the two factors $R_{\tilde\theta}$ and $R_I$. 
In this section, we aim to show that the regret bound (\ref{regretbound}) is order-optimal up to a logarithmic factor. In more detail, we offer a lower bound for $R_I$ by using some conventional methods in bandit theory and for $R_{\tilde\theta}$ we use Cramér-Rao lower bound.

In this section, we assume that the parameter $\theta$ is a scalar and as a result, we have some scalar variances $\sigma_i$ instead of the covariance matrices $\Sigma_i$. In addition, we need a mild assumption in our analysis. It is a property for the class of distributions which concludes a reverse Lipschitzness for the Fisher information with respect to the $\sigma$. To be more specific, let $I_{\sigma_1}^\theta$ and $I_{\sigma_2}^\theta$ be two Fisher information over $\theta$ by assuming $\sigma_1$ and $\sigma_2$ respectively as the values of variances. Then,
\begin{align} \label{IL}
    \vert I_{\sigma_1}^\theta - I_{\sigma_2}^\theta \vert \geq L \  \vert \sigma_1 - \sigma_2 \vert
\end{align}
To provide some intuition for this assumption, lets focus on Normal distribution as a compelling example,
\begin{align}
    \Pcal_{ \boldsymbol\theta} = \{ \Ncal (\theta , \sigma_1) | \theta \in \R \} \Rightarrow I_{\sigma_1}^\theta = \frac{1}{\sigma_1} \\
    \Pcal'_{ \boldsymbol\theta} = \{ \Ncal (\theta , \sigma_2) | \theta \in \R \} \Rightarrow I_{\sigma_2}^\theta = \frac{1}{\sigma_2}
\end{align}
Now recall the assumption we had in (\ref{GammaAssumption}), that results in,
\begin{align}
    \sigma_1 , \sigma_2 \leq \Gamma \Rightarrow \vert \frac{1}{\sigma_1} - \frac{1}{\sigma_2} \vert = \vert I_{\sigma_1}^\theta - I_{\sigma_2}^\theta \vert \geq \frac{1}{\Gamma ^ 2} \ \vert \sigma_1 - \sigma_2 \vert
\end{align}
And it is the reverse Lipcshitz assumption with $L = \frac{1}{\Gamma^2}$ for the Gaussian case.
After this illustration, let's state the main theorem of this section,
\begin{theorem} \label{LowerBound}
For every policy $\pi$ there is an environment $\nu$ such that $Reg(n) \in \Omega (n^{-\frac{2}{3}})$.
\end{theorem}
By environment $\nu$, we mean a configuration of arms' distributions. In more detail,
\begin{align}
    \nu = (\Pc_{\theta,\sigma_1}, \Pc_{\theta,\sigma_2},\dots, \Pc_{\theta,\sigma_k})
\end{align}

The proof of Theorem (\ref{LowerBound}) will be discussed in the following subsections (\ref{5.1sub}) and (\ref{5.2 sub}).
\subsection{Lower Bound for the Term $R_I(n)$} \label{5.1sub}
When we focus on the term $R_I(n)$, we simply cope with a stochastic multi-armed bandit problem in which the regret is not defined by the attaining rewards directly, but some second-level features of the arms ($I^\theta_{a_i}$) contribute to form the regret. This means that we need a more complicated policy to achieving the lower bound and on the other hand, intuitively the lower bounds of the classic MAB problem work for this second-level version of the problem. Because we have a lack of information in this version. Our proof for this part originates from the one in chapter 15 of \cite{lattimore2020bandit}. We first make the proof compatible with our setting and then generalize it to non-gaussian distributions.\\
The key idea in this part is well-known in the MAB problem. Based on every fixed policy $\pi$, we consider two complementary environments in which at least one of them eventuates in a $R_I(n)$ belongs to $\Omega(n^{-\frac{3}{2}})$.\\
In order to tackle this problem, we need the following definitions and lemmas.

\begin{mydef}
    let $\Pcb_{\pi,\nu}$ be a probability measure on an arbitrary measurable space $(\Omega, \Fc)$ and $A_1, X_1, \dots ,\\ A_n,X_n$ are the random variables corresponding to the actions and rewards, then we have
    \begin{align}
        \Pcb_{\pi,\nu}(a_1,x_1,\dots,a_n,x_n) = \prod_{t=1}^n \pi(a_t | a_1,x_1,\dots,a_{t-1},x_{t-1}) \Pc_{a_t}(x_t)
    \end{align}
\end{mydef}
Indeed, $\Pcb_{\pi,\nu}$ measures the probability of the whole events occurred, including both actions and rewards, up to step n.

\begin{lemma}\label{AL}
    [\cite{lattimore2020bandit}, page 196]Let a fixed policy $\pi$ and two environments $\nu$ and $\nu'$. Then after running $\pi$ on to these two environments we have $\Pcb_{\pi,\nu}$ and $\Pcb_{\pi,\nu'}$. then,
    \begin{align}
        \Dc_{KL}(\Pcb_{\pi,\nu}, \Pcb_{\pi,\nu'}) = \sum_{i=1}^k \E_\nu [T_i(n)]\Dc_{KL}(\Pc_{a_i}, \Pc'_{a_i})
    \end{align}
\end{lemma}

\begin{lemma} \label{BH}
    [Bretagnelle-Hubber Inequality, \cite{lattimore2020bandit}, chapter 14] 
    Let $P$ and $Q$ be probability measures on the same measurable space $(\Omega ,\Fc)$, and let $\Ac \in \Fc$ be an arbitrary event. Then,
    \begin{align}
        P(\Ac) + Q(\Ac^c) \geq \frac{1}{2} exp\big(-\Dc_{KL}(P,Q)\big)
    \end{align}
    where $\Ac^c = \Omega \backslash \Ac$ is the complement of $\Ac$.
\end{lemma}
First we introduce a new regret $R_I^{(-1)}(n,\nu)$ which originates form the definition of $R_I(n)$ and has a more convenient form to deal with,
\begin{align}
    R_I^{(-1)}(n,\nu)  = n I^\theta_{a^*} - \E_\nu [\sum_{t=1}^n I^\theta_{A_t}] = \sum_{i=1}^{k} \E_\nu[T_i(n)] \Lambda_i,
\end{align}

where $\Lambda_i = I_{a^*}^\theta - I^\theta_i$.
Now as we mentioned in the main idea, we need to introduce two environments $\nu$ and $\nu'$, hence let,
\begin{align}
    \nu = (\Pc_{\theta,\sigma_1}, \Pc_{\theta,\sigma},\dots,\Pc_{\theta,\sigma}),
\end{align}
where all the distributions from arm $2$ to $k$ are identical and $I_1^\theta$ is the greatest Fisher information in environment $\nu$.
It means the best arm is the first one. Also we denote other Fisher information as $I^\theta_c := I^\theta_2 = I^\theta_3 = \dots = I^\theta_k$. Then after applying policy $\pi$ to the environment $\nu$ , we seek for the Achilles' heel of the policy, in more details,
\begin{align}
    h= \argmin_{j\geq2} \E_\nu[T_j(n)]
\end{align}
where the expectation is over $\Pcb_{\pi,\nu}$. Since $\sum_{i=1}^n \E_\nu [T_i(n)] = n$, it holds that 
$\E_\nu [T_h(n)] \leq \frac{n}{k-1}$.\\
The second environment is,
\begin{align}
    \nu' = (\Pc_{\theta,\sigma_1}, \Pc_{\theta,\sigma},\dots, \Pc_{\theta,\sigma^{**}},\dots, \Pc_{\theta,\sigma}),
\end{align}
where specifically the distribution of the $h$th arm, $\Pc_{\theta,\sigma^{**}} := \Pc_{\theta,\sigma_h}$ and $I^\theta_{**} := I^\theta_h > I_1^\theta$. In other words, now in environment $\nu'$, arm $h$ is the best one. Note that the only difference between $\nu$ and $\nu'$ is on the entry $h$.\\
Then we have the following bound for the regret of the first environment $\nu$,
\begin{align}
    R_I^{(-1)}(n,\nu) \geq \frac{n}{2}  \Pcb_{\pi,\nu}[T_1(n) \leq \frac{n}{2}] (I^{\theta}_1 - I^{\theta}_{c})
\end{align}
because $\Pcb_{\pi,\nu}[T_1(n) \leq \frac{n}{2}] \leq \sum_{i=2}^n \E_\nu [T_i(n)]$. With a same argument we have,
\begin{align}
    R_I^{(-1)}(n,\nu') \geq \frac{n}{2} \Pcb_{\pi,\nu'}[T_1(n) > \frac{n}{2}] \  (I^{\theta}_{**} - I^{\theta}_1).
\end{align}
Now let $ \Lambda := I^{\theta}_{**} - I^{\theta}_1 = I^{\theta}_{1} - I^{\theta}_c$, by using lemma (\ref{BH}), it holds, 
\begin{align}
    R_I^{(-1)}(n,\nu) + R_I^{(-1)}(n,\nu') &\geq \frac{n\Lambda}{2} \big[\Pcb_{\pi,\nu}[T_1(n) \leq \frac{n}{2}] + \Pcb_{\pi,\nu'}[T_1(n) > \frac{n}{2}]\big] \\ &\geq \frac{n\Lambda}{4} exp(\Dc_{KL}(\Pcb_{\pi,\nu}, \Pcb_{\pi,\nu'}))
\end{align}
Then lemma (\ref{AL}) implies that,
\begin{align}
    R_I^{(-1)}(n,\nu) + R_I^{(-1)}(n,\nu') &\geq \frac{n\Lambda}{4} exp(\Dc_{KL}(\Pcb_{\pi,\nu}, \Pcb_{\pi,\nu'})) \\ &\geq \frac{n\Lambda}{4} exp(-\E_\nu[T_h(n)] \Dc_{KL}(\Pc_{\theta,\sigma_1}, \Pc_{\theta,\sigma^{**}})).
\end{align}
Let us define $\tilde\Lambda := \Dc_{KL}(\Pc_{\theta,\sigma_1}, \Pc_{\theta,\sigma^{**}}) \leq \I^{\sigma_1} (\sigma^{**} - \sigma_1)^2$ where the inequality can be shown by Taylor expansion. It is worthy to mention that $\I^{\sigma_1}$ is the Fisher information of the distribution with respect to parameter $\sigma$. Since $\E_\nu [T_h(n)] \leq \frac{n}{k-1}$ we have,
\begin{align} \label{errrr}
    R_I^{(-1)}(n,\nu) + R_I^{(-1)}(n,\nu') \geq \frac{n\Lambda}{4} exp(-\frac{n}{k-1} \tilde\Lambda)
\end{align}
Also from (\ref{IL}), it holds,
\begin{align}
    \tilde{\Lambda} \leq \I^{\sigma_1} (\sigma^{**} - \sigma_1)^2 \leq \I^{\sigma_1} (L)^{-2} \ (I^\theta_{**} - I^\theta_1)^2 = \I^{\sigma_1} (L)^{-2} \ \Lambda^2.
\end{align}
Thereby we rewrite the equation (\ref{errrr}) as follows,
\begin{align}
    R_I^{(-1)}(n,\nu) + R_I^{(-1)}(n,\nu') \geq \frac{n\Lambda}{4} exp(-\frac{n}{k-1} \tilde\Lambda) \geq \frac{n\Lambda}{4} exp(-\frac{n}{k-1} C_1 \Lambda^2)
\end{align}
where $C_1 := \I^{\sigma_1} (L)^{-2}$. By choosing $\Lambda = \sqrt{\frac{k-1}{n C_1}}$,
\begin{align} \label{Rminus}
    R_I^{(-1)}(n,\nu) + R_I^{(-1)}(n,\nu') \geq \frac{1}{4e} \sqrt{\frac{k-1}{C_1} n} = C_2 \sqrt{n}
\end{align}
Hence at least one of the environments $\nu$ and $\nu'$ has a regret belongs to $\Omega(\sqrt{n})$.
Now we have to show that this result implies $R_I(n) \in \Omega (n^{-\frac{3}{2}})$,

\begin{align}
    R_I(n) = \frac{1}{\E[\sum_{t=1}^n I^\theta_{\pi_t}]} - \frac{1}{n I^\theta_1} &= \frac{n I^\theta_1 - \E[\sum_{t=1}^n I^\theta_{\pi_t}]}{ n \E[\sum_{t=1}^n I^\theta_{\pi_t}] I^\theta_1} \\ &\geq \frac{C_2 n^{\frac{1}{2}}}{n \E[\sum_{t=1}^n I^\theta_{\pi_t}] I^\theta_1}             \label{eq22}
    \\ &\geq \frac{C_2 n^{\frac{1}{2}}}{n^2 (I^\theta_1)^2} \geq C_3 n^{-\frac{3}{2}}  \quad (C_3 = \frac{L\sqrt{k-1} }{4e \sqrt{I^{\sigma_1}}(I^{\theta}_1)^2})\label{eq23}
\end{align}
where the inequality (\ref{eq22}) is from (\ref{Rminus}) and the first inequality in (\ref{eq23}) holds because $\E[\sum_{t=1}^n I^\theta_{\pi_t}] \leq n I^\theta_1$.
Therefore, we obtain a lower-bound with $\Omega(n^{-\frac{3}{2}})$ for the term $R_I(n)$. Now we turn to the term $R_{\tilde\theta}(n)$.

\subsection{Lower Bound for the Term $R_{\tilde\theta}(n)$} \label{5.2 sub}
From Cramer-Rao inequality, for any unbiased estimator we have,
\begin{align}
    Var[\tilde\theta] \geq \frac{1}{I^\theta}
\end{align}
Hence, when we restrict our estimators to the unbiased estimators, we simply have a lower bound for the variance by Cramér-Rao inequality.
Besides, it is worthy to mention that we have two source of randomness in this problem. First, the randomness induced from the algorithm. In other words, there are some algorithms that randomly choose the arms and gradually change their underlying distributions for selecting arms. The second source of randomness comes from the distribution of arms. To be more specific, we always have a random set of data even in the case that we run a naive deterministic algorithm that always choose the first arm. Cramér-Rao inequality can present a valuable result after the realization of the first randomness, when we know the selected arms,

\begin{align}
    Var[\tilde\theta \vert A_1,A_2,\dots,A_n] \geq \frac{1}{\sum_{t=1}^n I^\theta_{A_t}}
\end{align}
Now if we use law of total variances and conditioning $\tilde\theta$ to the output of the arm selection process, we have,
\begin{align}
    Var[\tilde\theta] = \E\big[Var[\tilde\theta \vert A_1,\dots,A_n]\big] + Var \big[ \E[\tilde\theta \vert A_1,\dots,A_n]\big] &\geq \E\big[Var[\tilde\theta \vert A_1,\dots,A_n]\big] \\ &\geq \E\big[ \frac{1}{\sum_{t=1}^n I^\theta_{A_t}}\big] \\ &\geq
    \frac{1}{\E \big[ \sum_{t=1}^n I^\theta_{A_t} \big]}
\end{align}
where the last inequality is Jensen's inequality due to the convexity of $\frac{1}{x}$.
Therefore we have,
\begin{align} \label{Rthetatils}
    R_{\tilde\theta}(n) = Var(\tilde\theta) - \frac{1}{\E \big[ \sum_{t=1}^n I^\theta_{A_t} \big]} \geq 0
\end{align}
\subsection{Combining The Results}
After proposing these two lower-bounds for the terms $R_I(n)$ and $R_{\tilde\theta}(n)$, Theorem (\ref{LowerBound}) clearly holds. By combining (\ref{eq23}) and (\ref{Rthetatils}), we have,
\begin{align}
    Reg(n) = R_{\tilde\theta}(n) + R_I(n) \geq 0 + C_3 n^{-\frac{3}{2}} \quad \Rightarrow \quad Reg(n) \in \Omega(n^{-\frac{3}{2}})
\end{align}
\section{Discussion} \label{section6}
As we mentioned earlier, the novelty of this work is mainly in the problem definition and approach to the distributed estimation problem with a bandit theory point of view.\\
Therefore, there are many aspects in the setting which are worthy to spend more time on. Now we point out some of them in a nutshell.
\subsection{The Logarithmic Factor}
In the stochastic multi-armed bandit problem, it is known that the UCB algorithm can be modified to exactly order optimal algorithm. For better understanding, one might look at chapter 9 of \cite{lattimore2020bandit}. Since our algorithm has roots in UCB so it might be a good question of interest whether the same argument can be applied here or not, i.e. get rid of the logarithmic term in the theorem (\ref{order}).
\subsection{Impacts of Dimension $d$}
As one can see in Theorem \ref{order}, the impact of $d$ on the analyzed regret is linear. This brings a couple of interesting questions. Is there any algorithm that can outperform LCB with better dependency on $d$? Is there any multi-dimensional version of Theorem \ref{LowerBound} that can characterize the contribution of $d$ to the regret?
\subsection{Further Distribution Families}
We are optimistic that the result on the performance of the algorithm can be generalized to more complex probability distributions (i.e. non-Gaussian settings). As an intuition, one might look at section (\ref{section4.2}), the first term of the decomposition demonstrates the second-order convergence rate of Cramér-Rao bound. This is a quantity that has been under research in the statistics literature. For instance, one can look at this article \cite{efron1975defining} in which this quantity is investigated for exponential families. This is important to mention that going beyond the Gaussian case needs a more sophisticated approach for addressing the trade-off between the two terms of these decompositions. The second terms might be usually controlled by modification of the conventional algorithms in bandit theory, on the other hand, the first terms might require extra efforts to make the same order of regret in comparison to the second ones.


\section{Acknowledgment}
The authors are grateful to Prof. Efron for his valuable guidance, Prof. Alishahi, and Prof. Froughmand for helpful discussions and insights.

\bibliographystyle{alpha}
\bibliography{references} 

\appendix
\section{Proof of Algorithm}

\begin{lemma}\label{traceineq} \textbf{Trace concentration} \\
let $X_1,X_2,\dots,X_n \sim \Ncal(\mu ,\Sigma)$ and $Z = Tr(\Sigma) - Tr(\hat\Sigma) $ where $\Sigmah =\dfrac{1}{n-1} \sum_{j = 1}^{n} (X_j-\bar{X})(X_j-\bar{X})^T$ be the unbiased estimator for the   covariance matrix $\Sigma$ ,then

\begin{align}
  \Prob\left[ \vert Z \vert > \eps \Vert\Sigma\Vert_F \right] \leq 2 \exp\left(-\frac{(n-1) ~  }{8}\min\left(\eps^2,\epsilon.\frac{\|\Sigma\|_F}{\|\Sigma\|_2}\right)\right)  .
\end{align}
\end{lemma}

\begin{lemma} \label{Goodevent} \textbf{Regret Analysis} 
\end{lemma}
\proof
Our proof is based on bounding the probability of the following event for every $i$,
\begin{align}
    G_i = \{ tr_1 > \max_{t\in[n]}  LCB_1(t,\delta)\} \bigcap \{ LCB_1(u_i,\delta) > tr_1\}
\end{align}
This is a good event because if $G_i$ happens then arm $i$ would be chosen at most $u_i$ times, i.e. $T_i(n) \leq u_i$.
Now we prove that event $G_i$ occurs with high probability, or equivalently upper bound the probability of $\Prob [G_i^c]$,
\begin{align}
    G_i^c = \{tr_1 \leq \max_{t\in[n]}  LCB_1(t,\delta) \} \bigcup \{ LCB_1(u_i,\delta) \leq tr_1\}
\end{align}
Now we bound the probability of each term separately, for the first term we have,
\begin{align}
    \Prob[tr_1 \leq \max_{t\in[n]}  LCB_1(t,\delta) ] \leq \Prob[ \exists s : tr_1 \leq LCB_1(s,\delta)] \leq \sum_{s=1}^n \Prob[ tr_1 \leq LCB_1(s,\delta)] \leq n\delta
\end{align}
For the second term,
\begin{align}
    \Prob[ LCB_1(u_i,\delta) \leq tr_1 ] &= \Prob[\hat{tr}_i(u_i) - \sqrt{\frac{8 \log(\frac{2}{\delta})}{u_i - 1}} \ksig \leq tr_1]  \\ & = \Prob[ \Delta_i - \sqrt{\frac{8 \log(\frac{2}{\delta})}{u_i - 1}} \ksig \leq tr_i - \hat{tr}_i(u_i)] \\ & \leq 2 exp(-\frac{(u_i -1) \eta_i^2}{8 \ksig^2}) \quad \quad \text{(from lemma (\ref{traceineq}))} \label{ineq57}
\end{align}
where $\eta_i = \Delta_i - \sqrt{\frac{8 \log(\frac{2}{\delta})}{u_i - 1}} \ksig $ and $u_i = \ceil{ \frac{8  \ksig^2 \log(\frac{2}{\delta})}{(1-c)^2\Delta_i^2}} + 1$.
One should note that by considering these values for $\eta_i$ and $u_i$, the two inequalities $\eta_i \leq c \Delta_i$ and $c \Delta_i \leq \frac{\Vert \Sigma_i \Vert_F^2}{\Vert \Sigma_i \Vert_2}$ hold for an arbitrary $c \in [0,\frac{1}{d}]$.
The inequality $\eta_i \leq \frac{\Vert \Sigma_i \Vert_F^2}{\Vert \Sigma_i \Vert_2}$, guarantees that for the upper-bound (\ref{ineq57}) we were in the regime of $\epsilon^2$ in lemma (\ref{traceineq}).
After this illustration,
\begin{align}
    \Prob[ \Delta_i - \sqrt{\frac{8 \log(\frac{2}{\delta})}{u_i - 1}} \ksig &\leq tr_i - \hat{tr}_i(u_i)] \leq 2 exp(-\frac{(u_i -1) \eta_i^2}{8 \ksig^2}) \\ & \leq 2 exp(-\frac{(u_i -1) c^2 \Delta_i^2}{8 \ksig^2}) \leq 2 exp(- \frac{ \log(\frac{2}{\delta} )c^2}{(c-1)^2})
\end{align}
putting everything together,
\begin{align}
    \Prob[G_i^c] \leq n \delta + 2 exp(- \frac{ \log(\frac{2}{\delta} )c^2}{(c-1)^2}).
\end{align}
Then, we are ready to bound $\E [T_i(n)]$,
\begin{align}
    \E [T_i(n)] = \E [ \ind \{ G_i\}T_i(n)] + \E [ \ind \{ G_i^c \}T_i(n)] & \leq u_i + n \Prob[G_i^c] \\ & \leq u_i + n[n \delta + 2 exp(- \frac{ \log(\frac{2}{\delta} )c^2}{(c-1)^2})] ,
\end{align}
So by substituting $u_i$, we have, 
\begin{align}
     \E [T_i(n)]\leq
      \ceil{ \frac{8  \ksig^2 \log(\frac{2}{\delta})}{(1-c)^2\Delta_i^2}} + 1+  n[n \delta + 2 exp(- \frac{ \log(\frac{2}{\delta} )c^2}{(c-1)^2})].
\end{align}
Now we choose $\delta = \frac{2}{n^\alpha}$ where $\alpha = \max\{2,(d-1)^2\}$, Therefor we can rewrite the inequality as follows,
\begin{align}
    \E [T_i(n)]\leq \ceil{ \frac{8 \alpha  \ksig^2 \log(n)}{(1-c)^2\Delta_i^2}} + 1 + 2 + 2 n ^{1- \frac{\alpha c^2}{(1-c)^2}}.
\end{align}
If we choose our arbitrary $c \in [0,\frac{1}{d}]$ equals to $\frac{1}{1+\sqrt{\alpha}}$,
\begin{align}
    \E [T_i(n)] \leq \frac{\C_d \log(n)}{\Delta_i^2} + 5,
\end{align}
where $C_d = 8 \ksig^2 \Big(1+\sqrt{\max\{2,(d-1)^2\}}\Big)^2$.

\end{document}